\documentclass{article} % For LaTeX2e
\usepackage[preprint]{colm2026_conference}

\usepackage{microtype}
\usepackage{enumitem}
\usepackage{hyperref}
\usepackage{url}
\usepackage{booktabs}
\usepackage{afterpage}
\usepackage{amsmath}
\usepackage{amssymb}
\usepackage{textcomp}
\usepackage{algorithm}
\usepackage{algpseudocode}

\usepackage{cleveref}
\usepackage{tikz}
\usetikzlibrary{arrows.meta,positioning,fit,backgrounds,shapes.misc,calc}
\usepackage[svgnames,x11names,dvipsnames]{xcolor}
\tikzset{
  box/.style={
    draw=none, fill=yellow, rounded corners=10pt,
    inner xsep=12pt, inner ysep=9pt, font=\large
  },
  smallbox/.style={
    box, font=\normalsize, inner ysep=7pt
  },
  groupLabel/.style={font=\Large, text=black},
  arrow/.style={-Latex, line width=1.2pt, black},
  ghost/.style={draw=black, very thick, rounded corners=10pt}
}

\usepackage{lineno}

\definecolor{darkblue}{rgb}{0, 0, 0.5}
\hypersetup{colorlinks=true, citecolor=darkblue, linkcolor=darkblue, urlcolor=darkblue}

\title{AdMem: Advanced Memory for Task-solving Agents}

\author{
  Runzhe Wang \\
  Princeton University \\
  \texttt{runzhew@princeton.edu} \\\And
  Huilin Lu \\
  Amazon \\
  \texttt{huilinlu@amazon.com} \\\And
  Shengjie Liu \\
  Amazon \\
  \texttt{zycjlsj@amazon.com} \\\AND
  Li Dong \\
  Amazon \\
  \texttt{ldonga@amazon.com} \\\And
  Jason Zhu \\
  Arm \\
  \texttt{jason.zhu@arm.com} \\
}

\begin{document}

\ifcolmsubmission
\linenumbers
\fi

\maketitle

\begin{abstract}
Large Language Models (LLMs) show promise as tool-using agents but remain limited in long-horizon tasks that require remembering, organizing, and reusing knowledge. Prior memory approaches aim to resolve the situation, but mainly focus on storing factual information. Recent work on procedural memory improves task reuse, yet often reduces to replaying past successes without addressing failure cases or online scalability. We introduce a unified and automatic memory framework that integrates semantic, episodic, and procedural memory in a bi-level design combining short-term and long-term stores. A multi-agent architecture with actor, memory, and critic agents enables automatic memory generation, reward annotation, and adaptive retrieval. Long-term memory is managed through reward-based evaluation, merging, and pruning, ensuring scalability and continual improvement. Experiments across various environments show that our approach improves robustness and success on long multi-turn tasks compared to existing baselines. This work highlights the importance of comprehensive, adaptive memory for advancing LLM-based agents.
\end{abstract}

\section{Introduction}
Large Language Models (LLMs) (\cite{brown2020languagemodelsfewshotlearners, JMLR:v24:22-1144,deepseekai2025deepseekr1incentivizingreasoningcapability,touvron2023llama2openfoundation,zeng2023glm130bopenbilingualpretrained}) have driven major progress in artificial intelligence, achieving breakthroughs across many areas. While they have improved in reasoning and tool use within agentic settings, they still struggle with long multi-turn tasks that require remembering, organizing, and applying knowledge across sessions or large inputs \cite{zhang2024surveymemorymechanismlarge}. These challenges highlight the central role of memory, which is widely regarded as an essential component of intelligence and crucial for agent adaptivity. Two main research directions have emerged to tackle the memory challenge in agentic LLMs \cite{wu2025humanmemoryaimemory}: (1) \textit{architectural memory}, where additional capacity is built directly into the network (e.g., through layers, modules \cite{xu2023parameterefficientfinetuningmethodspretrained}), and (2) \textit{symbolic / textual memory} systems that the model reads and writes explicitly (often via providing APIs or policies). Across both directions, many efforts are made in an attempt to enhance the LLM's information handling capability in the long task solving process.

One line of work integrates memory directly into the model parameters to enhance the effective context length of the model. End-to-End Memory Networks \cite{sukhbaatar2015endtoendmemorynetworks} introduced differentiable attention over external memory for question answering, and later work \cite{bergesmemory, behrouz2024titanslearningmemorizetest} extend the idea to introduce learned long-term memory modules that retains historical context while keeping training parallelizable and inference efficient. Parametric approaches provide speed and differentiability but sacrifice readability and controllable persistence, since the remembered information is bound to model parameters and is neither interpretable nor controllable by designers.

In contrast, textual stores are auditable and tool-friendly, but require well-designed policies to be effective and efficient. Early works \cite{modarressi2025memllmfinetuningllmsuse, modarressi2024retllmgeneralreadwritememory} fine-tunes models to write knowledge as triples and to read via retrieval over that store, improving language modeling and knowledge-heavy tasks with interpretable memory traces; Memory Sandbox \cite{huang2023memorysandboxtransparentinteractive} foregrounds user control, letting human user add / merge / delete memories directly; MemGPT \cite{packer2024memgptllmsoperatingsystems} introduces OS-style virtual context management, with a focus on memory paging for the LLM to perform CRUD operations; Mem0 \cite{chhikara2025mem0buildingproductionreadyai} extracts memory units from running conversations, consolidates them (often into graphs), and reports major latency and token-cost wins in production-style settings; Mem1 \cite{zhou2025mem1learningsynergizememory} goes further, training a compact internal textual state that’s recurrently updated across turns; \cite{zhong2023memorybankenhancinglargelanguage} layers episodic histories, long-term summaries, and evolving user “portraits,” with decay/boost rules inspired by forgetting curves; merging systems like MemAgent \cite{yu2025memagentreshapinglongcontextllm} train a reinforcement-learned memory policy that reads long texts in segments and overwrites / consolidates memory to achieve linear-time long-context processing. HiAgent \citep{hu2025hiagent} decompose task-solving into multiple sub-task trunks for better context condensation.

While the above work shows miscellaneous designs on memory generation and management policies, they primarily build the memory in two considerations: 1. learn factual information about the world and the user; 2. condense the long history context into short and portable text segments useful for subsequent inferences. From the perspective of cognitive architectures \citep{cohen1997memory}, they revisit the classical division of semantic and episodic memory \cite{wu2025humanmemoryaimemory}. Meanwhile, an important part in the division is missing in the process as the procedural memory, with LLMs functioning as decision makers in a probabilistic production system and prompts serving as control flow. Therefore several recent work \cite{fang2025mempexploringagentprocedural,tang2025agent,wang2025mirixmultiagentmemoryllmbased,wang2024agentworkflowmemory,yang2024bufferthoughtsthoughtaugmentedreasoning} incorporate the procedural memory as a documentation of successful decision making process for future reference. Agent workflow memory (AWM) \cite{wang2024agentworkflowmemory} induces reusable workflows from past successes in web navigation. By storing and applying these workflows, AWM achieves notable improvements in success rates and efficiency on challenging benchmarks such as Mind2Web \cite{deng2023mind2webgeneralistagentweb} and WebArena \cite{zhou2024webarenarealisticwebenvironment}; building on this, Mem$^p$ \cite{fang2025mempexploringagentprocedural} develops a lifelong procedural memory that distills experiences into granular instructions and higher-level templates, dynamically updating and transferring them across tasks and models, with strong results on TravelPlanner \cite{xie2024travelplannerbenchmarkrealworldplanning} and ALFWorld \cite{shridhar2021alfworldaligningtextembodied}; beyond action trajectories, Buffer of Thoughts (BoT) \cite{yang2024bufferthoughtsthoughtaugmentedreasoning} focuses on reasoning patterns by storing distilled ‘thought templates’ to guide problem-solving, improving accuracy and robustness across reasoning tasks with minimal overhead; extending the scope further, MIRIX \cite{wang2025mirixmultiagentmemoryllmbased} introduces a multi-agent, multi-modal memory architecture with six structured types, enabling long-term, personalized, and efficient memory management across diverse benchmarks; Agent KB \cite{tang2025agent} advances cross-agent procedural memory, combining high-level strategies and execution logs in a hierarchical store to enable transfer across domains, achieving large gains on GAIA \cite{mialon2023gaiabenchmarkgeneralai} and SWE-bench \cite{jimenez2024swebenchlanguagemodelsresolve}.

However, these design of the procedural memory has some main limitations. First, they view the procedural memory as a recipe of instructions extracted from successful task completions, therefore the memory is not effective in guiding the critical task steps that the model usually fails at. Second, the binary task-level feedback considered in their setting is sparse and is not sufficient in guiding a long task solving process that involves many steps, a common challenge termed as credit assignment in reinforcement learning designs. Third, they mostly consider the memory generation in an offline training-inference setting, with evaluations built on Question-Answering-type datasets, and there is very limited scalable memory management design in online deployments, lacking in automatic memory entry evaluation, active erasing and reranking memory, and adaptive memory recall.

Theoretically, elegant and general-purpose agent frameworks are proposed \citep{sumers2023cognitive,
gao2025survey}. Two direction of thoughts shape an ideal agent with strong intelligence: a memory architecture perspective with short-term vs long-term memory, episodic, semantic and procedural memory; a learning cognitive perspective with a feedback loop and memory encoding/decoding as motor that propels an agent to self evolve through experience. However a gap persist between these ideas and practical memory implementations in the literature. Conceptual frameworks lack implementation on any practical LLM applications. In particular, there are substantial mismatches between rule-based, deterministic systems, statistics-based numerical algorithms, and
stochastic, token-level reasoning processes in LLMs.
Practical systems often simplify or omit critical components required for learning in interactive environments. Systems implementing learning memory typically lack mechanisms for reward-driven updates, belief modeling, transition learning, or credit assignment, limiting their ability to improve over time.

Given the above, we hope to design a better memory system for a life-long task-solving agentic environment, aiming towards bridging the aforementioned theory-practice gap. Our motivation for adding memory to LLM-based agents is (1) to enable user and task customization, and (2) to support self-improvement in the agent's decision making process in a long-run environment. In our work, we introduce AdMem, a unified framework of
\begin{itemize}[
    itemsep=0pt,
    topsep=0pt,
    parsep=0pt,
    partopsep=0pt]
   \item A comprehensive memory system that supports the generation, storage, management, and retrieval of procedural, semantic, and episodic memories. A bi-level memory architecture is established, from short-term memory designed for context compaction, to scalable long-term memory with automatic memory evaluation, consolidation and pruning, and adaptive memory retrieval.
   \item An agent planning paradigm designed for effective memory generation, incorporating adaptive task planning, expectation commenting, and automatic reflections.
\end{itemize}

\section{Methodology}
\subsection{Set up}
We consider the task-solving agentic setting where an agent aims to solve a series of tasks by interacting with an exterior environment in turns. The environment possibly comprises different components including human users, other agents or tool-calling infrastructures, and the interactions can be realized by natural language communication or LLM tool uses. In each round $t$, the agent takes action $a_t$ which is transferred to the environment. Then the environment evolves and responses with an observation $o_t$. The agent's goal, defined by each task, is to achieve certain state of the environment respectively with its actions. Notice that in real applications like human assistants, the environment usually does not reset after each task, making a life-long horizon for the problem, which is different from typical task solving benchmarks where different tasks utilize independent environment states.

We formalize the agent decision making process to also be a Partially-Observable Markov Decision Process (POMDP) $(S,A,T,R,\Omega)$. The agent preserves an agent state $s\in S$ which we also refer to as memory. In each round $t$, the agent  chooses action $a_t\sim \pi(s_{t})$ with an LLM-based policy $\pi(c_t)\in\Delta(A)$ that maps an LLM context ($c_t$, generated from memory $s_t$) to a probability distribution over the action space. Then it receives observation $o_t\in \Omega$ from the environment and performs memory update as state transition $s_{t+1} = T(s_{t}, a_t, o_t)$. 

Standard vanilla baselines for multi-turn task solving benchmarks (e.g. \citet{ma2024agentboard}) use a naive implementation of the POMDP to test the native decision making ability of LLM models, namely setting $c_t=s_t = [a_1,o_1,a_2,o_2,a_3,o_3\cdots a_t,o_t]$ to be the trajectory so far. Then an LLM model is used as a policy to generate the next action $a_{t+1}\sim \pi_{LLM}(c_{t})$. As the trajectory can be very long for complex tasks, these benchmarks raise challenges for modern LLMs to support and effectively reason over long contexts. Meanwhile, as an agentic approach to this challenge, prior work has proposed two main directions: (1) Context Compaction, maintaining the state $s_t$ at a moderate length that fits within a single LLM context window, typically through truncation or LLM summarization, and (2) Retrieval Augmented Generation, providing only a partial state to the LLM when generating actions. These two handling patterns naturally correspond to the short-term and long-term memory in human cognition.

\subsection{Framework}

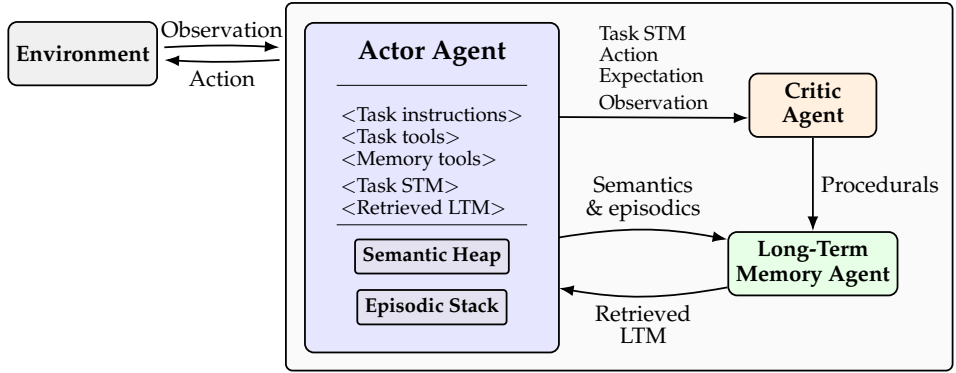
\begin{figure*}[htbp]
\centering
\resizebox{0.9\linewidth}{!}{
\begin{tikzpicture}[
  node distance=10mm and 20mm,
  title/.style   ={font=\bfseries\large},
  label/.style   ={font=\bfseries\small},
  small/.style   ={font=\small},
  tinytext/.style={font=\scriptsize},
  box/.style={draw, rounded corners, thick, minimum width=30mm, minimum height=10mm, align=center},
  >=Latex,
  envbox/.style   ={box, fill=gray!12},
  actorbox/.style ={box, fill=blue!10},
  criticbox/.style={box, fill=orange!12},
  membox/.style   ={box, fill=green!10},
  heap/.style     ={draw, rounded corners=1.5pt, thick, fill=BlueViolet!10},
  stack/.style    ={draw, rounded corners=1.5pt, thick,fill=BlueViolet!10},
  outer/.style    ={draw, rounded corners=3pt, thick, fill=black!2}
]

% Nodes
\node[envbox,font=\bfseries, minimum width=2cm] (user)    {Environment};

\draw[outer] (3.2,0.8) rectangle (13.7,-5);
\node[actorbox,anchor=north, minimum width=4cm,
  minimum height=5.2cm, align=center,
  anchor=north] at (5.5,0.5) (actor) {};
\node[title] at (5.5,0) {Actor Agent};
\node[criticbox,anchor=north, font=\bfseries, minimum width=2cm] at (11.5,-0.3) (critic) {Critic\\Agent};
\node[membox,anchor=north, font=\bfseries, minimum width=2cm] at (11.5,-2.8) (memory) {Long-Term\\Memory Agent};

\node[anchor=north, small] at (5.5, -0.7) {\shortstack[l]{\textless Task instructions\textgreater\\\textless Task tools\textgreater\\\textless Memory tools\textgreater\\\textless Task STM\textgreater \\\textless Retrieved LTM\textgreater}};

\draw[-] (4, -0.5) to (7,-0.5);
\draw[-] (4, -2.7) to (7,-2.7);
\node[heap, label] at (5.5, -3.2) {Semantic Heap};
\node[stack, label] at (5.5, -4.0) {Episodic Stack};

%coordinate
%\draw[step=1cm, gray!40] (-1,-5) grid (14,1);
%\draw[thick,->] (-1,0) -- (14.2,0) node[right] {$x$};
%\draw[thick,->] (0,-5) -- (0,1.2) node[above] {$y$};

\draw[->, thick, bend left=3] ($(user.east) + (0.1,0.1)$) to node[midway, above]{Observation} ($(3.2,0) + (-0.1,0.1)$); 
\draw[<-, thick, bend right=3] ($(user.east) + (0.1,-0.1)$) to node[midway, below]{Action} ($(3.2,0) + (-0.1,-0.1)$);

\draw[->, thick] ($(7.5, -1)$) to node[midway, above, font=\small]{\shortstack[l]{Task STM\\Action\\Expectation\\Observation}} ($(critic.west)+(0, -0.2)$);  
\draw[->, thick] (critic) to node[midway, right]{Procedurals} (memory);      
\draw[->, thick, bend left=10] ($(7.5, -2.9)$) to node[midway, above = 0mm]{\shortstack{Semantics\\\& episodics}} (memory);  
\draw[<-, thick, bend right=10] ($(7.5, -3.7)$) to node[midway, below]{\shortstack{Retrieved\\LTM}} (memory);      

\end{tikzpicture}}
\caption{Interaction diagram between the external environment, actor, critic, and memory agents. The actor agent maintains short-term memories (STM) to get all the necessary information about the task, and is provided with long-term memories (LTM) to utilize past experiences.}
\label{fig:interaction_diagram}
\end{figure*}
In our memory framework, we incorporate both short-term memory and long-term memory as our agentic memory. Specifically, our agent state $S=S_{\text{short}} \times S_{\text{long}}$ comprises the short term memory part $S_{\text{short}}$ and the long term memory part $S_{\text{long}}$. To enable efficient memory management, we build our system as a multi-agent framework that incorporates three important parts, an actor agent, a memory agent, and a critic agent.
\begin{itemize}[
    itemsep=0pt,
    topsep=0pt,
    parsep=0pt,
    partopsep=0pt]
    \item \textbf{Actor agent:} an LLM-based agent that utilizes memory to interact with the environment to solve tasks. The agent maintains a short-term memory state $s_t\in S_{\text{short}}$ in itself that is updated per turn and is cleared at the end of each task, where the information is encoded into the long term memory. We will refer this short term memory as state in this paper.
    \item \textbf{Long-term memory agent:} a retrieval-based agent that maintains the long-term memory store $M\in S_{\text{long}}$. It stores three types of memory: semantic, episodic and procedural memory that are used across turns and tasks.
    \item \textbf{Critic agent:} an LLM-based agent that helps the memory generation process. It aims to compress the raw memory material, annotate the procedural memories with guidance and reward signals so that the memories can help the actor agent to improve its later decision-making process.
\end{itemize}

We consider three types of information $M = M_S \oplus M_E \oplus M_P$ (the semantic, episodic and procedural memory) stored in memory to follow the classic taxonomy in cognitive science.

    \begin{itemize}[
    itemsep=0pt,
    topsep=0pt,
    parsep=0pt,
    partopsep=0pt]
    \item $\bf{M_S}$: semantic memory stores the facts and general knowledge about the environment and the tasks. 
    \item {$\bf{M_E}$:} episodic memory stores the summary of past events in the agent's interactions.
    \item $\bf{M_P}$: procedural memory stores the decision making guidance extracted from past actions. With each action made and outcome observed, the agent will think about whether the outcome meets its expectation and contributes to the task, and whether the action can be improved. It then encodes everything as a procedural memory entry that guides subsequent decision-making under similar circumstances.
\end{itemize}

The semantic memories and episodic memories are relatively easy to form as we use a native LLM to summarize facts and events. The procedural memories are more complicated as (1) we can only evaluate the action until all the related outcomes are observed, and (2) there may not be explicit reward provided by the environment. In our setting, the success of a solution attempt may depend on multiple steps, so we introduce a critic agent to temporarily cache action scenarios and conclude the action at a proper time later. Besides, when the agent generates an action, we ask it to explicit formulate the expected purpose or result of the action, and the critic agent will conclude the action by comparing the actual outcome with the expected outcome. The complete memory generation and usage pipeline is illustrated in the next section.

\subsection{Agent Pipeline}\label{sec:pipeline}

\begin{algorithm*}
\caption{Actor Agent Pipeline}
\label{alg:actor}
\begin{algorithmic}
\State Initial agent state $s_0$. Round number $t\leftarrow 0$. 
\While{active}
    \State $t\leftarrow t+1$. Receive observation $o_t$ from the user / environment.
    \State Report $o_t$ to the critic agent.
    \State Report the context $c_t = (s_{t-1}, o_t)$ to the memory agent to generate semantic and episodic memory.
    \State Call the memory agent to retrieve procedural memory $M_t = [m_t^{(1)},m_t^{(1)}\dots m_t^{(n)}]$, as well as semantic and episodic memories $M'_t$, given the context $c_t$.
    \State Choose action $a_t$ based on $(c_t, M_t, M'_t)$, and set up an expected outcome $\hat{o}_{t+1}$ for the action. Transit to state $s_t \leftarrow p(s_{t-1}, o_t, a_t)$ (namely update the short term memory).
    \State Report the half-baked memory $k = (c_t, a_t,\hat{o}_{t+1}, M_t)$ to the critic.
\EndWhile
\end{algorithmic}
\end{algorithm*}

\begin{algorithm*}
\caption{Critic Agent Pipeline}
\label{alg:critic}
\begin{algorithmic}
\State Memory Queue $Q\leftarrow []$.
\While {active}
    \If{received half-baked $k$ from the actor} 
        \State put $k$ into $Q$.
    \EndIf
    \If{received observation $o$ from the actor} 
        \For {every half-baked memory $k=(c_k,a_k,\hat{o}_{k},M_k)$ in Q}
            \If {$o$ is relevant to $k$}
                \State Append $o$ to the entry $k$.
                \If {$k$ has received all the observations for criticism}
                    \State Generate reward $r_k\in[0,1]$ and reflection $f_k$ for $k$.
                    \State Report a new procedural memory entry $(c_k,a_k,f_k)$ to the memory agent
                    \State (to be inserted into the memory store).
                    \State Report reward for the retrieved memories $(M_k,r_k)$ to the memory agent
                    \State (to be used to update the memory evaluation scores).
                \EndIf
            \EndIf
        \EndFor
    \EndIf    
\EndWhile
\end{algorithmic}
\end{algorithm*}

\Cref{fig:interaction_diagram} provides an overview of the information passing pipeline in the operation of the agent. The critic and memory agent backup the actor agent by supporting a long-term memory for the actor agent to refer to, and they work in parallel to solve tasks. \Cref{alg:actor} and \ref{alg:critic} provide a detailed description of the agent execution pipelines.

In the execution pipelines, the actor agent is responsible for handling communication with the user and the environment, when receiving memory provided by the long term memory agent. Beside talking to the user and the environment, it can also use a set of memory planning tools to generate plans and thoughts that are helpful for task solving and memory generation. The critic agent is responsible for the generation of procedural memories. It receives agent action context (the state and observation before the action), a current agent action, the agent's expected purpose or outcome of the action, and the actual outcome as the raw material for generation the procedural memory. Then it condenses the raw material to summarize the context and action, proposes better alternatives as well as labels the procedure with reward $r_t\in\{0,1\}$ by evaluating whether the outcome meets the expectation and is helpful for solving the task. To help the critic agent obtain enough observations for criticism, we designed several planning tools for the actor agent that relates to memory generation:

\begin{itemize}[
    itemsep=0pt,
    topsep=0pt,
    parsep=0pt,
    partopsep=0pt]
    \item \textbf{Plan for the new task}: the actor agent generate a plan first when solving a new task, and a procedural memory for the planning is generated at the end of the task when the critic agent received all the observations related to the plan.
    \item \textbf{Add sub-goal}: generate a sub-plan for a sub-goal of the task. 
    \item \textbf{Conclude sub-goal}: a procedural memory is generated for the sub-goal planning action, with everything between adding sub-goal and concluding sub-goal as its observations.
    \item \textbf{Think}: generate some thoughts in the short term memory.
    \item \textbf{Act}: act in the environment. A procedural memory for the action is generated after immediate observation.
\end{itemize}

Adding sub-goals and concluding sub-goals provide the agent with the ability to explore the environment like a search tree. Meanwhile, we can manage the context in a way similar to the stack in the execution of a computer program, and save context space with better scope management, as elaborated in the next section. 

Moreover, notice that each time the critic agent generate a reward, it also passes reward to the long-term memory store together with indices of the retrieved procedural memory entries that guide the action that induces the reward. This reward is used for the memory agent to evaluate the effectiveness of each procedural memory to guide downstream tasks, as we illustrate in the next section.

\subsection{Short term memory for context compaction}
For the actor agent, an efficient context management is needed for the agent to focus on the current goal and carry out the current plan; for the critic agent, an efficient summarization of a plan execution is also needed. For context compaction, we adopted the idea from traditional computer science that a program maintains a running stack in its memory space. In our agent, the current stack includes the current sub-goal to be attained as well as a plan made earlier by the agent. Turns outside the sub-goal are compressed and summarized by the actor agent LLM. This stack-based structure allows the agent to maintain a clear execution context, attend only to information relevant to the current step, and resume higher-level objectives once lower-level sub-goals are completed.
\subsection{Reward-based long-term memory management}

The memory agent is a storage that manages long-term memories in our framework. It is responsible for long-term memory retrieval and management.
\begin{itemize}[
    itemsep=0pt,
    topsep=0pt,
    parsep=0pt,
    partopsep=0pt]
    \item \textbf{Memory retrieval:}  for episodic and semantic memory, we use dense retrieval to enable retrieval with similarity match. For procedural memory, we incorporate both the evaluation of memory entry effectiveness and the context similarity into consideration for retrieval. The details are explained later.
    \item \textbf{Memory management:} the agent prunes and evicts redundant memories to avoid excess occupancy. We use an LLM to manage all the semantic memories for information updates. For the procedural memories, we set a threshold $\epsilon>0$ for the disuse of memories that has stayed long enough in the store. Any memory retrieved at a frequency below $\epsilon$ are deleted from the store, and if any two memory entries are constantly retrieved or not retrieved together with frequency above $1-\epsilon$, the agent will merge the two memory entry into one, enabling a more effective retrieval process.
\end{itemize}

As we can see from above, the retrieval process is central to the memory management as the managements are built upon the retrieving statistics. Therefore, to enabling a more effective retrieval process, we wish to incorporate a bandit-type evaluation of the memory entries especially for procedural memories that directly link with reward information. Therefore, we build a simple model to estimate the effectiveness of each procedural memory recipe in guiding downstream tasks. 

For each procedural memory entry $m\in M_P$ in the store, we set up an adaptive parameter $v_m\in [0,1]$ indicating whether the suggestion in the entry is effective. In an actor step, the actor agent makes the decision in the face of context $c_t$, as well as retrieved procedural memories $M_t = [m_t^{(1)},m_t^{(1)}\dots m_t^{(n)}]$, as illustrated in \Cref{alg:actor}. Then after the actor receives reward $r_t\in \{0,1\}$, we build a lightweight model by assuming that the reward is produced through a binary stochastic process :$r_t|_{c_t} = l_t^{(1)} \vee l_t^{(2)} \vee \cdots \vee l_t^{(n)}\vee l_t.$
Here $l_t^{(i)}\in\{0,1\}$ is a random variable indicating whether memory entry $i$ helps the action given context, and $l_t\in\{0,1\}$ indicates whether the action is successful without the help of any memory. For a memory entry to be helpful, we assume that the its content should be effective in its own context, and also the context in the memory entry and the current task context should be similar. Therefore we have model:

$$l_t^{(i)} \sim \mathcal{B}(v_{m_t^{(i)}}\psi(c_t)^\top \psi(c^{(i)})).$$

Here $\mathcal{B}(p)$ is the Bernoulli distribution with mean $p$, $\psi$ is the normalized dense retrieval embedding map used in our memory retriever, and $c^{(i)}$ is the context of the action in memory entry $m_t^{(i)}$. Therefore, each time we observe a reward $r_t$, we can infer an update to the estimation of $v_{m_t^{(1)}}\cdots v_{m_t^{(n)}}$ in our model, and given the estimations of $v_m$, we can retrieve the most helpful memory entries greedily by selecting those with top-$n$ scores $v_m \psi(c_t)^\top \psi(c_m)$.

In practice we use an Expectation-Maximization (EM) algorithm to update our estimate of the $v_m$ parameters. In the spirit of optimism in the face of uncertainty, we set the initial $v_m$ values to 1.

\section{Experiments}

\afterpage{
\begin{table*}[h]
\centering\resizebox{1.0\linewidth}{!}{
\begin{tabular}{c|cc|cc|cc|cc}
\toprule
Domain  &  \multicolumn{2}{c|}{AlfWorld} &  \multicolumn{2}{c|}{Babyai} &  \multicolumn{2}{c|}{Jericho} &  \multicolumn{2}{c}{PDDL}  \\ \midrule
Number of tasks &  \multicolumn{2}{c|}{134} &  \multicolumn{2}{c|}{112} &  \multicolumn{2}{c|}{20} &  \multicolumn{2}{c}{60} \\
\midrule
Metric & C & P & C & P & C & P & C & P\\
\midrule
ReAct  & 49.3\%  & 0.6586 & 97.3\% & 0.9732 & 40.0\% & 0.4002 & 33.3\% & 0.5537\\ 
AWM & 47.0\%  & 0.5467 & 93.7\% & 0.93682 & 10.0\% & 0.2227 &43.3\% & 0.6407\\ 
\textbf{AdMem}  & \textbf{63.4\%}  & \textbf{0.7755} &\textbf{100.0\%} & \textbf{1.0000} &\textbf{50.0\%}& \textbf{0.5325}&\textbf{76.7\%} & \textbf{0.9069}\\\hline
\midrule
Domain  & \multicolumn{2}{c|}{\shortstack[l]{Tool-query\\Academic}} & \multicolumn{2}{c|}{\shortstack[l]{Tool-query\\Weather}}  & \multicolumn{2}{c|}{WebShop} & \multicolumn{2}{c}{Science World} \\
\midrule
Number of tasks & \multicolumn{2}{c|}{20} & \multicolumn{2}{c|}{20} & \multicolumn{2}{c|}{251} & \multicolumn{2}{c}{90} \\ 
\midrule
Metric & C & P & C & P & C & P & C & P\\
\midrule
ReAct  & 70.0\%  & 0.8365 & 60.0\% & 0.8565 & \textbf{42.2\%} & 0.7507 & \textbf{58.9\%} & 0.8258\\ 
AWM & 55.0\%  & 0.8185 & 25.0\% & 0.7529 & 29.2\% & 0.7012 & 31.1\% & 0.6324\\ 
\textbf{AdMem}  & 
\textbf{94.7\%}  & \textbf{0.9671} & \textbf{75.0\%} & \textbf{0.9036}& 41.4\% & \textbf{0.7645} & 56.7\% & \textbf{0.8447}\\
\bottomrule
\end{tabular}}

\caption{Task completeness (C) and average progress (P) on AgentBoard, using Claude Haiku 4.5.}
\label{tab:results2}
\end{table*}

\begin{table*}[h]
\centering
\begin{tabular}{c|cc|cc|cc}
\toprule
Domain  & \multicolumn{2}{c|}{Jericho}  & \multicolumn{2}{c|}{\shortstack[l]{Jericho\\Second epoch}} & \multicolumn{2}{c}{\shortstack[l]{Jericho\\Third epoch}} \\ 
\midrule
Metric & C & P & C & P & C & P \\
\midrule
LLM + React &  40.0\%  & 0.4002&  - & - & - & - \\ 
LLM + STP & 40.0\%& 0.4553&  - & - & - & - \\ 
LLM + LTP  & 20.0\% & 0.4257& 25.0\% & 0.4775 & 30.0\% & 0.4743\\
LLM + AdMem &  \textbf{45.0\%}& \textbf{0.5569}& 50.0\% & 0.6578& \textbf{60.0\%}& \textbf{0.6820}\\ \bottomrule
\end{tabular}

\caption{Task completeness (C) and average progress (P) on AgentBoard, using Claude Haiku 4.5. For the ablation tests, 1. we remove long-term memory recalls and tested only short-term planning (STP); 2. We remove short-term planning and context management, and tested vanilla LLM with per-step long-term procedural memory (LTP). 3. We keep all the elements in AdMem. For part 2 and 3, we further ask the agent to re-test all the tasks in each additional epoch without resetting the long-term memory.} 
\label{tab:results}
\end{table*}
}
We evaluates our memory mechanisms under streaming multi-task conditions that simulates real task-solving agent environments. AdMem is tested on a variety of task domains bundled by the implementation of AgentBoard environment \citep{ma2024agentboard}. The benchmark provides a variety of tasks over several domains ranging from embodied AIs, text games, web research and tool calling. In each domain, we set up the agent with the same system prompts (\Cref{sec:appendix2}) and the vanilla task instructions in AgentBoard, and launch a life-long run over the set of all tasks once, while accumulating and using memory in an online sense. We tested the task completion over the run.

In building our agent, we adopted Claude Haiku 4.5 model as our LLM backbone. We baseline our methodology against existing memory frameworks including Agent pipelines without across-task memory (ReAct, \citet{yao2022react}) and Memory-based agents for procedural knowledge (AWM, \citet{wang2024agentworkflowmemory}). Our current baseline selection (AWM) is intended to represent agents with lifelong memory accumulation in multi-turn task settings, while many existing memory systems (e.g., Mem0, MemGPT, MIRIX) primarily focus on retrieval optimization over static knowledge bases, which are well-suited for question-answering tasks but not directly designed for interactive tool-use or long-horizon decision-making environments. The result is exhibited in \Cref{tab:results2} with both task completeness (the percentage of task completed) and the average progress ($p_i\in[0,1]$ for each task $i$, $p_i=0$ for no progress and $p_i=1$ for completion). AdMem has been shown to boost the model performance in most domains while maintaining on-par best performances in the others.

The benefit of memory usually depends on the domain and the task sets because they vary with the amount of transferable knowledge and experience across different tasks. For previous procedural memory implementations (AWM in \Cref{tab:results2}) or our naive implementation of procedural memory (LLM + LTP in \Cref{tab:results}), we often observe that adding memory harms the performance when the transferred information causes more confusion than assistance across tasks. This actually manifests the challenge to distinguish task-only vs. transferable knowledge, as well as balancing memory-reliance vs. improvising. With the use of planning tools and careful memory encoding, \Cref{tab:results2} shows that: on the domains where memory is useful, our implementation can boost performance; on the domains where memory may not be useful, our implementation does not greatly hurt performance with the additional structure.

To further understand the effect of different components in our build, we decompose the system into (i) the acting and planning component (including short-term memory management), and (ii) the long-term memory component. Specifically, we evaluate performance when removing (a) procedural memory together with the reward model, and (b) planning and semantic-episodic memory encoding. Furthermore, we also test AdMem when the task set is streamed for multiple epochs, to see if the memory incur benefits over longer horizons. The result is captured in \Cref{tab:results}. Short-term planning, supported by memory planning tools and context management, contributes to the task progress but does not scale with the horizon as no information is shared across tasks. Long-term memory, on the other hand, genuinely improves task-performance over long horizons when similar situations and tasks are revisited, and thus is central to the self-evolution of the agent itself. However, when added alone, it burns-in with early performance degradation. 

Our design intentionally prioritizes task performance, learning capability and robustness in long-horizon settings over minimal per-step cost. Meanwhile, in the implementation of AdMem, the actor, critic, and memory agent work in parallel through multi-threading, so memory encoding and management do not incur much time delay to the whole pipeline. It is still a limitation for the current pipeline to incur extra prompts, planning steps, and a number of extra background LLM callings in solving the tasks. For the worst case in the above environments, per-step time cost is 2 times that for vanilla LLM with 3 times LLM calling, which happens when the task is simple so vanilla LLM is extremely fast.

\section{Conclusion}
In this work, we present AdMem, a systematic memory design for task-solving agents, incorporating a short-to-long term hierarchy a splitting of semantic, episodic, and procedural components. This design not only supports task customization but also enhances agent performance by enabling learning from past experiences. Our framework demonstrates strong effectiveness in the simulated environments in AgentBoard, particularly when agents face repetitive tasks. Furthermore, it demonstrates the potential for agents to autonomously evolve via memory-based self-reflection in online environments, paving the way for scalable and sustained long-term development.

\bibliography{colm2026_conference}
\bibliographystyle{colm2026_conference}
\appendix

\section{LLM Prompts}
\label{sec:appendix2}
Here we include the prompts used by the actor and the critic agent. The bold trunks are filled with corresponding information in the task-solving process.\\

\fbox{
  \parbox{0.9\linewidth}{\large \textbf{Actor Agent:}\\\noindent\rule{\linewidth}{0.4pt}\\
\small You are a helpful agent that solves tasks in an given environment. You take turns to interact with the environment and learn about the environment from the observations, until you achieve the goal defined in the task. A log for your previous turns in the task is given below, where planner messages (if exists) marks the plan you made earlier for solving the task and agent\_summary messages provide summarization of one of more turns in the process. \\\\In the next turn you MUST use one of the tools in the \textless tools\textgreater tag. Do not output plain text. Some tools are provided for you to plan for next steps in the task and some tools are provided to interact with the environment. You need to decide which tool to use, but to solve hard task efficiently, careful planning is important. Always make sure you generate valid JSON for using the tool. \\\\Before action, the system will also generate some messages by recalling the system memory. These messages include information from your past interactions with the environment, as well as reflections on your past actions under similar circumstances (for other tasks in the same environment). Please examine these messages and think about their implications for the possible outcomes or your action, before deciding your action.\\\\\textless tools\textgreater\\\hspace*{2em}\textbf{[Memory Tool Descriptions] }\\\textless/tools\textgreater\\\\Below is the description of the task and the environment.\\\\
\hspace*{2em}\textbf{[Task Instructions]}\\
\\Below is the task execution log:\\\\\hspace*{2em}\textbf{[Short Term Memory Output]}\\\\
Below is some memory recall for past experience under similar circumstances:
\\\\\hspace*{2em}\textbf{[Long Term Memory Output]}\\}}
\newpage
\fbox{
  \parbox{0.9\linewidth}{\large \textbf{Critic Agent:}\\\noindent\rule{\linewidth}{0.4pt}\\
\small You are a helpful assistant that generates a procedural memory entry from an event. The event is an action (or a plan) taken by an agent when solving a task. The purpose of the procedural memory entry is to provide a description of the context of the action (plan), the action (plan) taken by the agent, the expected outcome of the action (plan), a summary of the actual outcome after the action (plan), and some evaluations and reflections aiming to optimize the action based on the result. The procedural memory should not be too long, but should be helpful in improving the agent's behavior when provided to the agent in a similar situation in the future. Hide private information like user name, account number etc. to protect privacy, for example use [user name] in the place of actual user name. \\\\The context of the action (or plan):\textbf{[context]}\\\\The action (or plan) taken by the agent:\textbf{[action]}\\\\The expected outcome/purpose:\textbf{[expectation]}\\\\The actual outcome:\textbf{[observation]}\\\\
Below is some domain information that helps you understand the task setting.\\
\\
Now it's your turn to generate the procedural memory entry. To generate, call the tool 'generate\_procedural\_memory' and fill in the arguments with corresponding information.
\\\\\textless tool\_description\textgreater \\
\textit{Name:} generate\_procedural\_memory\\
\textit{Description:} Generate procedural memory entry.\\
\textit{Args:}\\
\hspace*{1em}\textit{context (str):} the action context, a brief and high-level background description summarizing the context where the agent action occurs. \\
\hspace*{1em}\textit{result (str):} a summary of the result of the action or plan, including 1. the agent's expected outcome / achievement; 2. the actual outcome; 3. whether they are consistent.\\
\hspace*{1em}\textit{success (bool):} True or False that the action was successful, considering 1. whether the outcome of the action meets the expectation, 2. whether the outcome offers progress towards the task solving and plan execution.\\
\hspace*{1em}\textit{reflection (str):} Think about what can be learned from the result of the action, specifically what to change in the agent's expectation, and what should the agent do under the same circumstances to better carry out the plan and solve the task. Specifically, think about whether the agent should set up a subgoal first instead of directly interacting with the environment, to better organize the task solving process.\\ \textless /tool\_description\textgreater}\\}\\

We also provide the descriptions for the memory tools we built. Notice that not all memory tools are activated simultaneously to the agent: at the start of a task, the agent can only use plan\_for\_the\_task. Later it can use add\_subgoal if the stack depth of sub-goals is within limit; it can use conclude\_subgoal if there is an active sub-goal; it can use think if the previous turn is an action in the environment, and it can use take\_action always after task planning.\\

\fbox{
  \parbox{0.9\linewidth}{\large \textbf{Memory Tools:}\\\noindent\rule{\linewidth}{0.4pt}\\
  \small \textit{Name:} plan\_for\_the\_task\\
  \textit{Description:} Plan for a solution. Set the goal and highlight the key points in solving the task, and plan for the task by writing a list of step-by-step instructions. Each instruction can either be high-level subgoal (e.g. learn something about the environment or achieve certain status), or some actions to interact with the environment (e.g. use certain tools in the environment or communicate with the environment). If the task is hard, decompose the task into multiple subgoals that are needed to be attained in line to solve the task.\\
\textit{Args:}\\
\hspace*{1em}\textit{goal (str):}The goal and key points of the current task.\\
\hspace*{1em}\textit{plan (str):} the step-by-step plan for solving the task.\\\noindent\rule{\linewidth}{0.4pt}

\textit{Name:} add\_subgoal\\
\textit{Description:} Add the a sub-goal to the current task and make plans to achieve the sub-goal. The sub-goal should admit an observation-based success criterion and should be helpful to solving the final task. Use this method when the task is hard to complete or when you believe the current task plan is not effective to solve the task any more.\\        
\textit{Args:}\\
\hspace*{1em}\textit{subgoal (str):} The current sub-goal to be set.\\
\hspace*{1em}\textit{plan (str):} A list of step-by-step instructions that can be followed to achieve the sub-goal.\\\noindent\rule{\linewidth}{0.4pt}
\textit{Name:} conclude\_subgoal\\
\textit{Description:} Conclude the current sub-goal judge whether it is successfully achieved. You should call the method if the current sub-goal is already achieved or it cannot be achieved anymore. \\
\textit{Args:}\\
\hspace*{1em}\textit{evaluation (str):} A short natural language justification of whether the sub-goal is successful\\
\hspace*{1em}\textit{success (bool):} Whether the sub-goal is successfully achieved.
\\\noindent\rule{\linewidth}{0.4pt}
\textit{Name:} take\_action\\
\textit{Description:} Take an action in the environment. In this method, you can take an action in the environment by pass the action information in the action argument. Write what you expect the action to achieve (or the purpose of the action) in the expectation argument.\\
\textit{Args:}\\
\hspace*{1em}\textit{action (str):} The action to be taken.\\
\hspace*{1em}\textit{expectation (str):} The expected outcome / purpose of the action.
\\\noindent\rule{\linewidth}{0.4pt}
\textit{Name:} think\\
\textit{Description:} Generate some thoughts. This method will not change the environment.\\
\textit{Args:}\\
\hspace*{1em}\textit{thoughts (str):} Some thoughts.\\}}

\end{document}